# Improving circuit miniaturization and its efficiency using Rough Set Theory

Sarvesh S S Rawat[1], Dheeraj Dilip Mor[1]
[1] School of Electronics and Instrumentation
VIT University
Vellore, India
sss.sarvesh888@gmail.com
mordheeraj91@gmail.com

Sanjiban Sekhar Roy[2], Anugrah Kumar[2]
Rohit kumar[2]
[2] School of Computing Science and Engineering
VIT University
sanjibanroy09@gmail.com
anugrah18@gmail.com
rrohit992@yahoo.co.in

*Abstract-* High-speed, accuracy, meticulousness and quick response are notion of the vital necessities for modern digital world. An efficient electronic circuit unswervingly affects the maneuver of the whole system. Different tools are required to unravel different types of engineering tribulations. Improving the efficiency, accuracy and low power consumption in an electronic circuit is always been a bottle neck problem. So the need of circuit miniaturization is always there. It saves a lot of time and power that is wasted in switching of gates, the wiring-crises is reduced, cross-sectional area of chip is reduced, the number of transistors that can implemented in chip is multiplied many folds. Therefore to trounce with this problem we have proposed an Artificial intelligence (AI) based approach that make use of Rough Set Theory for its implementation. Theory of rough set has been proposed by Z Pawlak in the year 1982. Rough set theory is a new mathematical tool which deals with uncertainty and vagueness. Decisions can be generated using rough set theory by reducing the unwanted and superfluous data. We have condensed the number of gates without upsetting the productivity of the given circuit. This paper proposes an approach with the help of rough set theory which basically lessens the number of gates in the circuit, based on decision rules.

## I. INTRODUCTION

In 1982, Z Pawlak [1] revolutionized the world with rough set theory to deal with uncertainty and vagueness present in ambiguous data [4]. It is now being implanted in numerous domains for realistic purposes which include knowledge discovery, pattern matching, data acquisition, expert systems, process control and machine learning [6]. The process of extracting decision relies on available data to analyze features, and also the generation of classification rules is done without the requirement of any previous information [4][3]. It provides a mathematical approach to eradicate the superfluous data i.e. getting the minimal set of rules having the same knowledge as the original one; being able to find the hidden pattern [2]. It originates from set theory; classifying objects based upon the attributes. Rough set theory is based on the classification of the universe of discourse induced by the indiscernibility relation, and is being successfully used in attribute reduction. There are many fields in which rough set theory is actively used as a helping and processing tool

## II. Rough Set Theory: Basic Concepts

An information system consists of all the information $IS = (U, W, V, f)$, data flowing through the circuits that can be represented by set of finite objects called universe $U$, $W = (w_1, w_2, w_3 \ldots)$ is a finite set of attributes, in our case they are set of wires, $V_w$ is the domain of attribute $W$. Each object in the universe is represented by a vector. $f : U \times W \to V$ is a function such that $f(x, w) \in V_w$ for every $w \in W, x \in U$ is called an information function. In our case U is set of 16 possible cases, W is set of 13 attributes, while D is decision attributes i.e., final bit from the OR gate.

## III. Approximation And Reduction

Let us assume that P is an equivalence relation based on U. The approximation space will be a pair $(U, P)$, where $U = (x_1, x_2, x_3 \ldots)$ and $X$ is an element such that $X \in U$. So we have two subsets that are the upper and lower approximations Z Pawlak [1].

$\underline{P}X = \{x \mid [x] \subseteq X\}$

$\overline{P}X = \{x \mid [x] \cap X \neq \emptyset\}$

The boundary condition is represented by $(\overline{P}X - \underline{P}X)$.

These sets are known as rough sets of $X$. Also called as lower ($\underline{P}X$) and upper approximation ($\overline{P}X$) spaces of $X$.

## IV. Indiscernibility In Circuit

The fundamentality of rough set theory insists the notion of indiscernibility. If a binary relation $P \subseteq X \times X$ is reflexive, transitive and symmetric then it is an equivalence relation. Indiscernibility holds such a relation. The equivalence relation $[X]_p$ of an element contains all the objects $y \in X$, such that $xRy$. Let us take $IS = (U, W)$ be an information system, then with any $C \subseteq W$

$IND(C) = \{(x, x') \in U^2 \mid \forall a \in C, a(x) = a(x')\}$

Where $IND(C)$ is called C- indiscernibility and $(x, x') \in IND(C)$, If $(x, x') \in IND(C)$ then object $x$ and $x'$ and indiscernible to each other. Equivalence class of C – indiscernibility relation denoted by $[x]_c$ informally two objects are indiscernible if one object cannot be distinguished from the other on the basis of a given set of attributes. Hence, indiscernibility is a function of the set of attributes under consideration [6]. An indiscernibility relation partitions the set of facts related to a set of objects into a number of equivalence classes. An equivalence class of a particular object is simply the collection of those objects that are indiscernible to the object it is often possible that some of the attributes or some of the attribute values are superfluous. This enables us to discard functionally redundant information.

## V. Core and reducts

The proposed approach basically reduces the number of gates in the circuit, based on smart algorithm. The reducts of an information system is said to be a reduced set of information table denoted by RED(R). The "core" is intersection of all sets [1][7]. So it has some sort of relation sets that are contained in all reducts, and is denoted by CORE(R). Mathematically [3][5] it can be represented as

$CORE(R) = \cap RED(D)$.

A decision table can have as many as *reducts*. As our primary concern is to deal with the smartest and best circuit miniaturization, we are not required to find each one of them. The reducts having minimal number of attributes is selected. Figure 1 shows the general process extracting the knowledge from data.

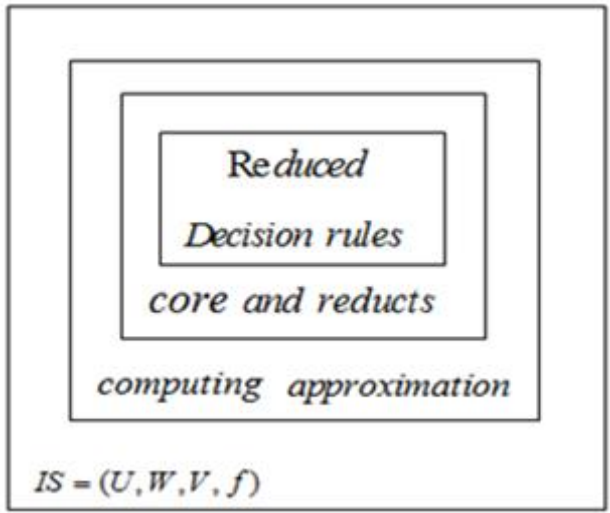

Figure 1. Extraction of knowledge using Rough Set

Now, we can readily use the above formulized ideas in the miniaturizing of the circuits.

## VI. Case Study

In this section, a simple structure using logic gates is shown in figure 2, which is the magnified view of a portion of complicated circuit and it is further reduced using Rough Set theory which relies upon logical classifier and the rules .

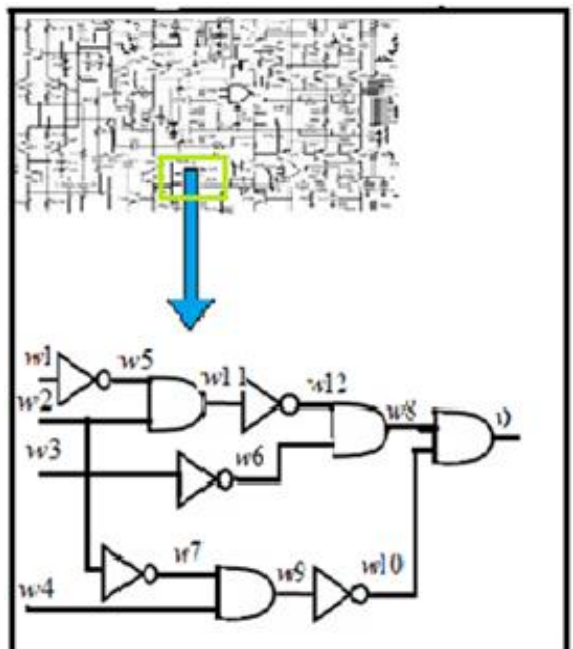

*Figure 2. Enlarge view of a circuit*

The Rough Set based approach is illustrated as follows. Here the different cases are represented in terms of attributes. This is the most widely used approach to extracting meaningful information. In our example, a set of data is generated by each gate in binary form (either 0 or 1), and the wires represents that attributes. The basic idea behind circuit miniaturization is to mine the data that is obtained as a result of each gate in a logical manner using algebraic developments so that the final result is not altered. This process can be easily done using rough set technique. The example that is shown here is a small circuit but the same technique can be implemented in bigger circuits using the same procedure. The data table of the circuit is shown below in table 1.

**Table 1**

| Attributes | | | | | | | | | | | | D |
|---|---|---|---|---|---|---|---|---|---|---|---|---|
| w1 | w2 | w3 | w4 | w5 | w6 | w7 | w8 | w9 | w10 | w11 | w12 | D |
| 0 | 0 | 0 | 0 | 1 | 1 | 1 | 1 | 0 | 1 | 0 | 1 | 1 |
| 0 | 0 | 0 | 1 | 1 | 1 | 1 | 1 | 1 | 0 | 0 | 1 | 0 |
| 0 | 0 | 1 | 0 | 1 | 0 | 1 | 0 | 0 | 1 | 0 | 1 | 0 |
| 0 | 0 | 1 | 1 | 1 | 0 | 1 | 0 | 1 | 0 | 0 | 1 | 0 |
| 0 | 1 | 0 | 0 | 1 | 1 | 0 | 0 | 0 | 1 | 1 | 0 | 0 |
| 0 | 1 | 0 | 1 | 1 | 1 | 0 | 0 | 0 | 1 | 1 | 0 | 0 |
| 0 | 1 | 1 | 0 | 1 | 0 | 0 | 0 | 0 | 1 | 1 | 0 | 0 |
| 0 | 1 | 1 | 1 | 1 | 0 | 0 | 0 | 0 | 1 | 1 | 0 | 0 |
| 1 | 0 | 0 | 0 | 0 | 1 | 1 | 1 | 0 | 1 | 0 | 1 | 1 |
| 1 | 0 | 0 | 1 | 0 | 1 | 1 | 1 | 1 | 0 | 0 | 1 | 0 |
| 1 | 0 | 1 | 0 | 0 | 0 | 1 | 0 | 0 | 1 | 0 | 1 | 0 |
| 1 | 1 | 0 | 0 | 0 | 1 | 0 | 1 | 0 | 1 | 0 | 1 | 1 |
| 1 | 1 | 0 | 1 | 0 | 1 | 0 | 1 | 0 | 1 | 0 | 1 | 1 |
| 1 | 1 | 1 | 0 | 0 | 0 | 0 | 0 | 0 | 1 | 0 | 1 | 0 |
| 1 | 1 | 1 | 1 | 0 | 0 | 0 | 0 | 0 | 1 | 0 | 1 | 0 |

## VII. Approximations

From our information system, we have two classes of decision set as 0 and 1. As the data value is discrete (either 0 or 1) so the total number of lower approximations is equal to that of the upper approximations. The accuracy of approximation is given by $\varphi_p(X) = \frac{|(\underline{P}X)|}{|(\overline{P}X)|}$. So

the quality of both the approximation is given by the equation which is

$$\gamma_p(X) = \frac{|U - (\bigcup_{t \in \{2,...,n\}} Bn_p(x_t))|}{|U|}$$

On solving the equation it is coming out to be 1.00 as $Bn_p(x) = 0$. The lower and upper approximations are shown in the table 2.

Table 2

| s.no. | Classes | Number of objects | Lower = upper approximation | Accuracy |
|---|---|---|---|---|
| 1 | 0 | 11 | 11 | 1.00 |
| 2 | 1 | 4 | 4 | 1.00 |

### VIII. Successive Reduction Technique

In general, for a given decision table many reducts may exist. When the information table is large it becomes very difficult to perform the task. With ROSE2 all 44 reducts are derived as in pairs {8, 10},{6, 10, 12},{3, 10, 12},{8, 9},{6, 9, 12},{3, 9, 12},{4, 7, 8},{2, 4, 8}{6, 10, 11},{3, 10, 11},{4, 6, 7, 12},{3, 4, 7, 12},{5, 6, 7, 10},{1, 6, 7, 10},{6, 9, 11}{3, 9, 11},{2, 4, 6, 12},{2, 3, 4, 12},{2, 5, 6, 10},{1, 2, 6, 10},{3, 5, 7, 10},{1, 3, 7, 1,0} {5, 6, 7, 9},{1, 6, 7, 9},{4, 6, 7, 11},{3, 4, 7, 11},{2, 3, 5, 10},{1, 2, 3, 10},{2, 5, 6, 9}{1, 2, 6, 9},{3, 5, 7, 9},{1, 3, 7, 9},{4, 5, 6, 7},{1, 4, 6, 7},{2, 4, 6, 11},{2, 3, 4, 11}{2, 3, 5, 9},{1, 2, 3, 9},{3, 4, 5, 7},{1, 3, 4, 7},{2, 4, 5, 6},{1, 2, 4, 6},[2, 3, 4, 5},{1, 2, 3, 4} . As the decision is fixed in this information table (either 1 or 0), we have to further reduce the data sets. Presently we have 44 reducts with different strengths. We will consider only those pairs that have strength of 100 %. On further reduction, we end up with 10 final reducts based on which our improved circuit will be drawn. They are {8,9}, {8,10}, {3,9,11,}, {3,9,12}, {6,9,11},{6,9,12},{6,10,12}{6,10,11}.{3,10,11}{3,10,12}.

### IX. Decision Rules:

On the basis of the above reducts, the following decision rules can be drawn from the data table

Table 3

| S.no. | IF(condition) | Then(decision) |
|---|---|---|
| 1. | (w3 =0)&(w9=0) &(w11=0) | D=1 otherwise 0 |
| 2. | (w3=0)&(w10=1) &(w12=1) | D=1 otherwise 0 |
| 3. | (w3=0)&(w9=0) &(w12=1) | D=1 otherwise 0 |
| 4. | (w6=1)&(w9=0) &(w12=1) | D=1 otherwise 0 |
| 5. | (W8=1)&(w9=0) | D=1 otherwise 0 |
| 6. | (w6=1)&(w10=1) &(w12=1) | D=1 otherwise 0 |
| 7. | (w6=1)&(w10=1) &(w11=0) | D=1 otherwise 0 |
| 8. | (w8=1)&(w10=1) | D=1 otherwise 0 |
| 9. | (w6=1)&(w9=0) &(w11=0) | D=1 otherwise 0 |
| 10 | (w3 =0)&(w10=1) &( w11=0) | D=1 otherwise 0 |

From these rules, we can simplify the complicated circuit more effectively .As an example we have taken Rule 1, to proposed a miniaturized structure of the given circuit.

### Conclusion:

Improving the efficiency, accuracy and low power consumption in an electronic Circuit is always been a bottle neck problem. So the need of circuit miniaturization is always there. To cope up with this problem we have anticipated an Artificial intelligence (AI) based approach that uses Rough Set Theory for its implementation. We have reduced the number of gates without affecting the output of the given circuit using the mathematical model of Rough Set Theory. It saves a lot of time and power that is wasted in switching of gates , the wiring-crises is reduced, cross-sectional area of chip is reduced, the number of transistors that can implemented in chip is multiplied many folds. Moreover, operational efficiency and effectiveness of chip is also improved.